\documentclass[runningheads]{llncs}
\usepackage{graphicx}
\usepackage{comment}
\usepackage{amsmath,amssymb} % define this before the line numbering.
\usepackage{color}
\usepackage{multirow}
\usepackage[width=122mm,left=12mm,paperwidth=146mm,height=193mm,top=12mm,paperheight=217mm]{geometry}

\begin{document}
% \linenumbers
\pagestyle{headings}
\mainmatter
\def\ECCVSubNumber{}  % Insert your submission number here

\title{ParaCNN: Visual Paragraph Generation via Adversarial Twin Contextual CNNs} % Replace with your title

%\begin{comment}
%\titlerunning{ECCV-20 submission ID \ECCVSubNumber}
%\authorrunning{ECCV-20 submission ID \ECCVSubNumber}
\author{Shiyang Yan, Yang Hua, Neil Robertson}
\institute{Queen's University Belfast}
%\end{comment}
%******************

\begin{comment}
\titlerunning{Abbreviated paper title}
\author{First Author\inst{1}\orcidID{0000-1111-2222-3333} \and
Second Author\inst{2,3}\orcidID{1111-2222-3333-4444} \and
Third Author\inst{3}\orcidID{2222--3333-4444-5555}}
%
\authorrunning{F. Author et al.}
\institute{Princeton University, Princeton NJ 08544, USA \and
Springer Heidelberg, Tiergartenstr. 17, 69121 Heidelberg, Germany
\email{lncs@springer.com}\\
\url{http://www.springer.com/gp/computer-science/lncs} \and
ABC Institute, Rupert-Karls-University Heidelberg, Heidelberg, Germany\\
\email{\{abc,lncs\}@uni-heidelberg.de}}
\end{comment}
%******************
\maketitle

\begin{abstract}
Image description generation plays an important role in many real-world applications, such as image retrieval, automatic navigation, and disabled people support. A well-developed task of image description generation is image captioning, which usually generates a short captioning sentence and thus neglects many of fine-grained properties, e.g., the information of subtle objects and their relationships. In this paper, we study the visual paragraph generation, which can describe the image with a long paragraph containing rich details. Previous research often generates the paragraph via a hierarchical Recurrent Neural Network (RNN)-like model, which has complex memorising, forgetting and coupling mechanism. Instead, we propose a novel pure CNN model, ParaCNN, to generate visual paragraph using hierarchical CNN architecture with contextual information between sentences within one paragraph. The ParaCNN can generate an arbitrary length of a paragraph, which is more applicable in many real-world applications. Furthermore, to enable the ParaCNN to model paragraph comprehensively, we also propose an adversarial twin net training scheme. During training, we force the forwarding network's hidden features to be close to that of the backwards network by using adversarial training. During testing, we only use the forwarding network, which already includes the knowledge of the backwards network, to generate a paragraph. We conduct extensive experiments on the Stanford Visual Paragraph dataset and achieve state-of-the-art performance.
\end{abstract}

\section{Introduction}
\begin{figure}
    \centering
    \includegraphics[width=0.8\linewidth]{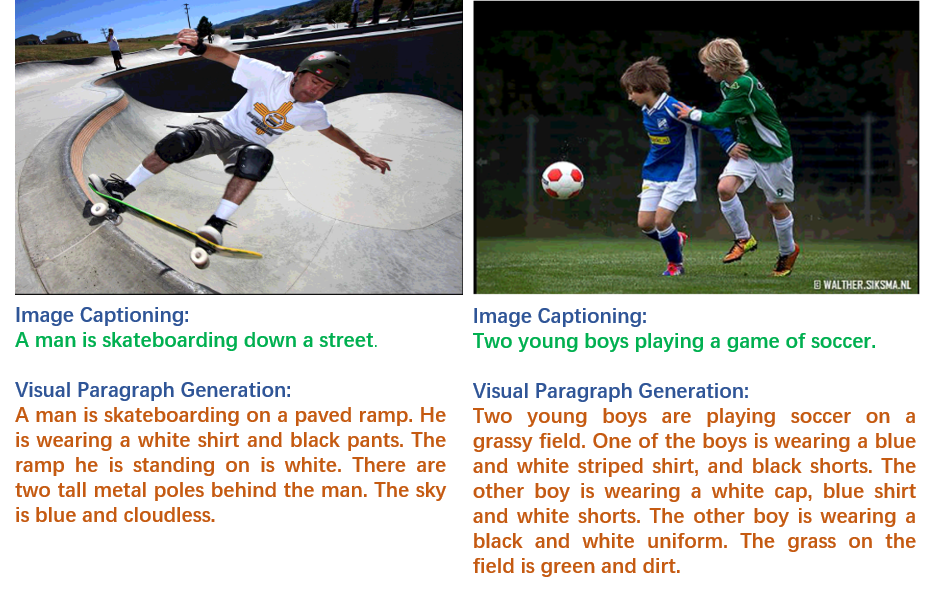}
    \caption{An illustration of the differences between image captioning and visual paragraph generation. The caption is a simplified sentence whilst the paragraph contains more fine-grained properties.  }
    \label{img:illustration}
\end{figure}

Image description generation~\cite{anderson2018bottom,aneja2018convolutional,kulkarni2013babytalk,cho2015describing} is a key topic in both the computer vision (CV) and natural language processing (NLP) communities. A classical task of image description generation is image captioning, which requires the machine to generate a caption to describe the image. Despite the encouraging progress~\cite{chen2017sca,anderson2018bottom,aneja2018convolutional} in image captioning, a single short sentence with less than twenty words is not enough to describe the full content of an image, which could be very informative on the subtle objects and their relationships. Furthermore, image captioning tends to capture the scene-level clues rather than fine-grained entities, limiting their direct applications in many real-world problems such as disabled people support and visual semantic navigation. Consequently, it is more natural to describe images in detailed paragraph, which have been studied recently~\cite{krause2016paragraphs,liang2017recurrent,chatterjee2018diverse,wang2018look,rennie2017self,luo2019curiosity}. An illustration of the differences between the image captioning and visual paragraph generation are shown in Figure~\ref{img:illustration}.

The challenges in visual paragraph generation are twofold: First, the sentences within one paragraph should be coherent and consistent. Secondly, encoding the long-term dependencies in the current sequence model, such as Recurrent Neural Networks (RNNs)~\cite{hochreiter1997long,cho2014learning}, is still challenging. To address these challenges, most of the previous research~\cite{krause2016paragraphs,liang2017recurrent,chatterjee2018diverse,wang2018look} use RNNs to form an encoder-decoder framework, often with hierarchical architecture to model the words and paragraph. The designed coupling mechanism between sentences in a hierarchical model~\cite{chatterjee2018diverse,liang2017recurrent} provides consistency and coherence within one paragraph. However, RNNs have a complex memorising, forgetting and addressing mechanism. Although Convolutional Neural Networks (CNNs) are usually not considered as the mainstream for the sequence to sequence tasks, recently, we see several improvements of CNNs in sequence to sequence tasks~\cite{gehring2017convolutional,aneja2018convolutional}, with encouraging results. As pointed out in~\cite{gehring2017convolutional}, compared to RNNs, computations in CNNs can be fully parallelised, and more importantly, the number of non-linearities in CNNs are fixed and independent to the input length, which is easier to be optimised. Nevertheless, the vanilla deep CNN model is not suitable for the visual paragraph generation, as it models the entire paragraph equally, which consequently neglects the coherence and consistencies between sentences within a paragraph.

In this paper, we propose a novel CNN architecture, ParaCNN, to effectively and efficiently tackle the task of visual paragraph generation. This architecture contains a hierarchy of topic convolutions and word convolutions. We first use the CNN network to decompose the visual features into several visual topics. Next, the new generated visual topic, combined with the contextual features from the previously generated sentence, is applied in the word convolutions~\cite{aneja2018convolutional} to generate a new sequence of words. This hierarchical structure solves the problem of coherence by using the context as the coupling mechanism between sentences within a paragraph. The parallel computing capability of the pure CNN model is more efficient than an hierarchical RNN structure and proves to be better in modelling long sequences. Unlike the hierarchical RNN-based  model~\cite{krause2016paragraphs,liang2017recurrent,chatterjee2018diverse,wang2018look}, our CNN model can be flexible enough to generate variable length of paragraphs, which is a very appealing property in real-world applications.
%The model is denoted as ParaCNN.

As argued in~\cite{serdyuk2017twin}, sequence to sequence model should form a better summarising of the past. However, the training objective of current sequence model corresponds with one-step prediction, where the local correlations are usually stronger than long-term dependencies and thus dominate the learning. Hence, inspired by the `twin' training scheme~\cite{serdyuk2017twin}, we propose an adversarial twin net training scheme. We force the hidden features of the forwarding network to be close to that of the backwards one by using adversarial training~\cite{goodfellow2014generative}. We only use the forwarding network in inference. The twin structure can force the ParaCNN to learn a good summarising of the past by making the model focus on the information that is useful to predict a specific token, which is also useful to the backward network. Our contributions can be summarised as follows: 1) We propose a visual paragraph model based on pure convolution operation, denoted as ParaCNN. The ParaCNN is flexible to generate a various number of sentences, which is more suitable in real-world applications. To our best knowledge, we are the first to use pure convolutions to generate a visual paragraph. 2) We propose an adversarial twin training scheme for the ParaCNN, which significantly improves performance. 3) We perform extensive experiments to validate the proposed methods and provide insights for future studies.

\section{Related Works}
\subsection{Image Captioning}
Most of the existing image captioning models use a deep CNN as the image embedding and feed the visual features to an RNN (LSTM/GRU) network. Since the attention mechanism shows improving performance on various NLP tasks~\cite{vaswani2017attention,bahdanau2014neural,yang2016hierarchical,golub2016character}, the visual attention mechanism~\cite{xu2015show,anderson2018bottom} has been extensively studied in image captioning. For instance, a channel-wise attention mechanism is proposed in~\cite{chen2017sca}. Semantic attention, which includes the image attributes, yields better results~\cite{you2016image}. Encoding fine-grained features from an object detector are studied ~\cite{anderson2018bottom,li2017image}. We find the visual attention mechanism~\cite{xu2015show} also improves the performance of the proposed ParaCNN. The features from an object detector can boost the performance since an object detector focuses on the semantically meaningful and fine-grained image features. In this paper, we also use object features \cite{johnson2016densecap} as the image embedding. The convolutional sequence processing for image captioning achieved comparable results with RNN-based approaches, and with much higher efficiency~\cite{aneja2018convolutional} because of a parallel training scheme. Other merits of utilising a CNN in image captioning include that the CNN can more easily handle long-term dependence~\cite{fan2018hierarchical} and avoids the complex memory and forgetting mechanism. The convolutional captioning model proposed in~\cite{aneja2018convolutional} inspires us to propose the ParaCNN model.

\subsection{Visual Paragraph Generation}
Regions-Hierarchical~\cite{krause2016paragraphs} introduces the first large-scale paragraph captioning dataset, which utilises the images from Visual Genome dataset and adds new paragraph annotations. The dataset shows more pronouns, verbs and more diversities than single sentence captioning dataset, which is more challenging. Regions-Hierarchical~\cite{krause2016paragraphs} proposes a hierarchical RNN model to generate the visual paragraphs. The hierarchical decoder includes two LSTMs, where the output of one sentence-level LSTM is used as input of the other word-level LSTM. Subsequent research uses a similar hierarchical approach. For instance, RTT-GAN~\cite{liang2017recurrent} extends this model with another paragraph-level LSTM and applied adversarial training. Diverse (VAE)~\cite{chatterjee2018diverse} also uses a hierarchical RNN structure with a more complex coupling mechanism between sentences and includes variational auto-encoder structure~\cite{kingma2013auto}, to diversify the generated visual paragraph.

It is worth mentioning is that a different approach~\cite{melas2018training} which uses a flat RNN to model the long paragraph. While their baseline model, which only uses a single RNN model yields poor performance, they achieve state-of-the-art results by introducing a repetition penalty sampling (Rep. Penalty Sampling) technique in self-critical training~\cite{rennie2017self}. We also apply the Rep. Penalty Sampling in the inference, but only use it in the testing stage. As a result, it requires little additional computation resource. The self-critical training in the long sequence is very computationally intensive, which makes it less practical in the real world. Our primary focus is not on the sampling method but the improvement of the model. Other methods like~\cite{wang2018look} use external knowledge of depth image to enrich the visual representation and achieve improved results.

\section{Methods}
We first introduce the problem formation of the visual paragraph generation, the training and inference mechanism of the proposed ParaCNN. Subsequently, we present the twin net training algorithm for the ParaCNN.

\subsection{ParaCNN}
\subsubsection{\textbf{Problem Formulation.}}

In image paragraph generation, we are given an image $I$ and are required to generate a paragraph of sentences $P = \{S_1, ..., S_M\}$. Each sentence $S$ contains $\{W_1, ..., W_N\}$ words. This structure of paragraph forms a hierarchical architecture which requires the model to generate a sequence of topics corresponding to the sentences, denoted as $T=\{T_1, ..., T_M\}$. Then the model should be able to generate the word sequence based on the generated topics. Next, we introduce the proposed ParaCNN by splitting into the training part and inference part.

\begin{figure*}[t]
    \centering
    \includegraphics[width=\textwidth]{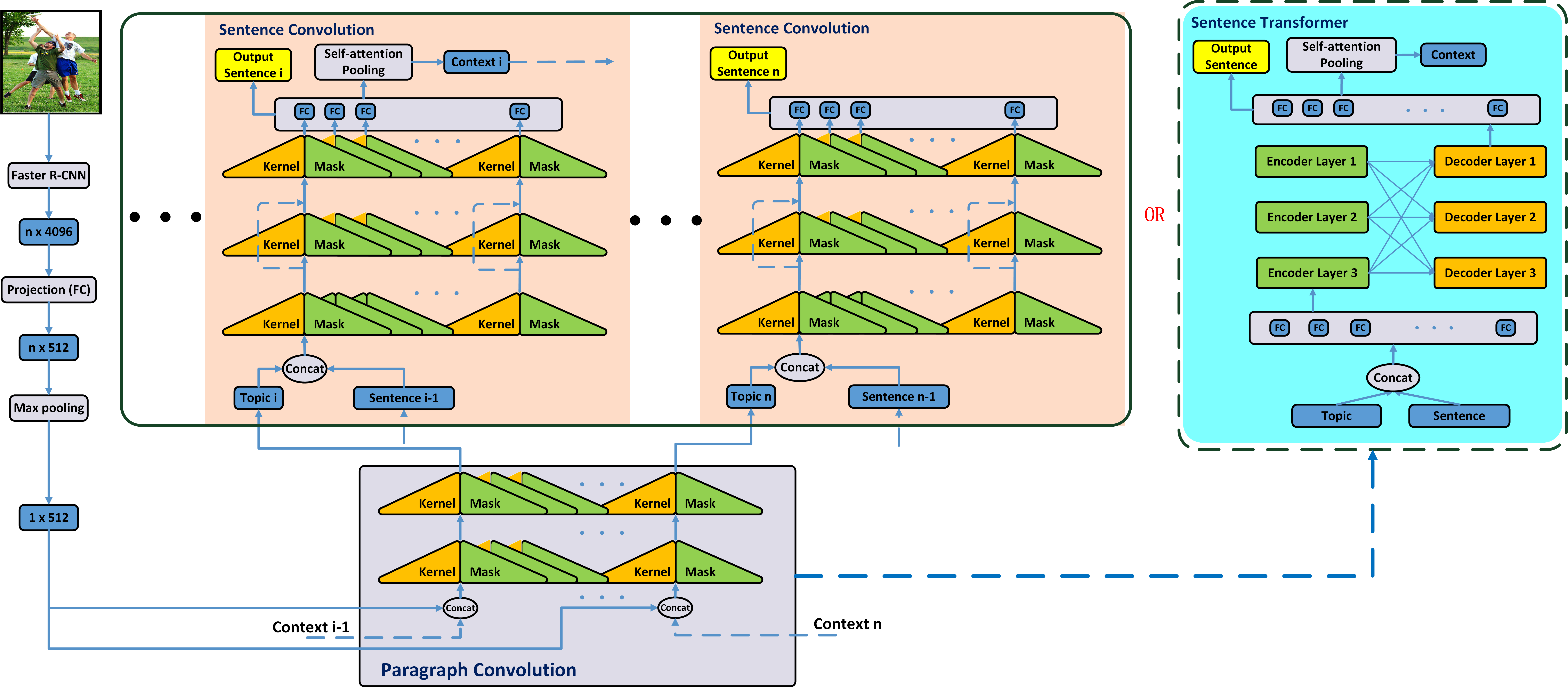}
    \caption{Illustration of ParaCNN: Each paragraph annotation is split into 6 sentences, which contains a maximum of 30 words. We generate new topic by using convolutional layers with the contextual information of the previously generated sentence; The pooling method for context generation can be configured as self-attention pooling or mean pooling. Then a sentence CNN model decodes the topic to a sequence of words. The whole network is computed in parallel and optimised with backpropagation. This figure is an example of illustration; actually, we use deeper CNN networks. Furthermore, the sentence generator (in pink) can be replaced with a transformer-based model (in blue), which proves the wide-range feasibility. \textit{Best viewed in color.}}
    \label{img:system}
\end{figure*}

\subsubsection{\textbf{Training.}}
We propose the contextual hierarchical CNN based on the above problem formulation, which is shown in Figure~\ref{img:system}. Firstly, we feed the image features into a CNN model, denoted as the paragraph convolution $f_p$, to generate $M$ topics. Each topic is then used to generate a sentence by using the convolutional sentence decoder $f_s$, similar to the one in~\cite{aneja2018convolutional}. Both the convolutional operations in paragraph convolution and sentence convolution are realised by applying the masked 1d convolution to make the model only look at the past information since we only have the past information during inference. More formally, the dynamics of the ParaCNN can be illustrated as follows:
\begin{small}
    \begin{equation}\label{sent}
        P_i(w_i^j|w_{<i}) = f_s(w_i^j; w_{<i}, T_{j})   \ \
        i= 1, ..., N; j=1, ..., M.
    \end{equation}
\end{small}\noindent It indicates the probability of generating the next token based on all the previously generated tokens are sampled from sentence convolutions $f_s$.
\vspace{-0.18cm}
\begin{small}
\begin{equation}\label{para}
        \begin{split}
            & S_{j-1} = {w_i^{j-1}, ..., w_N^{j-1}}  \\
            & context_{j} = Average(Self{\text -}attention(Embedding(S_{j-1}))),  \\
            & T_j = f_p(T_j; T_{<j}, I, context_{j}) \ \ { j=1, ..., M.}\\
        \end{split} 
\end{equation}
\end{small}\noindent where $S$ is the sentence, $context$ is the average of the Multi-head Self-attention~\cite{vaswani2017attention} pooled sentence embedding and $T$ is the topic vector, which is obtained from paragraph convolutions $f_p$. 

Equation~\ref{sent} provides a formal presentation of the word generation with convolutional decoder. The model generates a new word based on all the past information instead of the one in the last time step as in most of the RNN-based approaches. Equation~\ref{para} gives an illustration of the topic generation process. Specifically, the model generates a new topic based on the visual representations and the mean pooled embedding vector of the previously generated sentence.

If we consider the whole model as $f$, we can write down the model's dynamics as in Equation~\ref{whole}:
\begin{small}
    \begin{equation}\label{whole}
        P_i(w_i|w_{<i}) = f(w_i; w_{<i}, I) \ \ { i= 1, ..., M*N.}
    \end{equation}
\end{small}
\vspace{-0.6cm}
\subsubsection{\textbf{Inference.}}
The inference is performed sequentially, one word at a time. We first feed the start token `$<start>$' as $w_0$ to the model and then the $w_1 \sim p_1(w_1|w_0, I)$ is sampled. In this paper, we use greedy sampling in inference. Afterwards, we sequentially feed the generated tokens $w_1, ..., w_{i-1}$ to the model, and sample the generated new token $w_i \sim p_i(w_i|w_{<i}, I), i = 1, ..., M*N$.

\subsection{Twin ParaCNN}
\begin{figure*}[t]
    \centering
    \includegraphics[width=\linewidth]{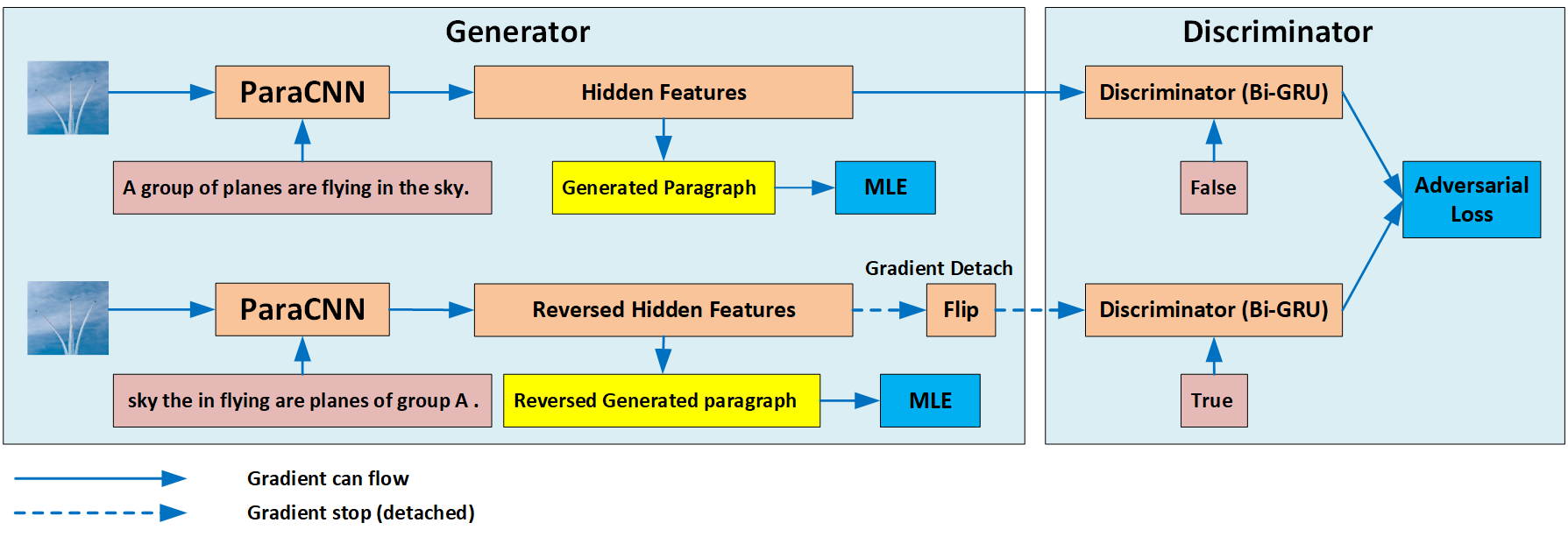}
    \caption{The twin network training pipeline with adversarial training in which the probability distribution of hidden features from the forwarding network is pulled close to that of the backwards network. \textit{Best viewed in color.}}
    \label{img:twin-gan}
\end{figure*}

\subsubsection{\textbf{Adversarial Twin Net.}}
Although the ParaCNN performs parallel computation and optimisation, its inference is still sequential. As discussed in~\cite{serdyuk2017twin}, the RNN-based sequence model cannot reason on the future information, which inspires them to propose the $TwinNet$ to force the RNN also focusing on the future tokens. In~\cite{serdyuk2017twin}, they use an L2 loss to force the distance between the hidden states of forwarding and backwards networks to be close, if they have the same ground-truths.

However, L2 loss cannot model the distribution; therefore, we propose to use the adversarial training~\cite{goodfellow2014generative} to push the distribution of the forwarding and backwards ParaCNN close to each other. We define the forwarding network as a general training process using MLE. The backwards network is implemented by flipping the ground-truth labels and feeding them to the model. Hence the output hidden features contain backwards information. Specifically, we use a more stable Wasserstein-GAN (W-GAN)~\cite{arjovsky2017wasserstein} to deploy the adversarial training framework. During the discriminator training, we set the hidden feature of the forwarding network as false while the hidden features of the backwards network as true to ensure the forwarding network to generate indistinguishable hidden features of the backwards network. During the training of the forwarding network, we get the probabilities of the hidden features being true and set them as the adversarial loss to fine-tune the forwarding network. The training process is presented in Figure~\ref{img:twin-gan}.

\subsubsection{\textbf{Advantages of the adversarial twin model.}}
Empirically speaking, the adversarial twin model's advantages are twofold:
First, if the model needs to predict $w_i$ based on $w_0, ..., w_{i-1}$, the twin networks learn the transition knowledge not only on $w_{i-2}$ to $w_{i-1}$ but also on $w_{i-1}$ to $w_{i-2}$, which makes the transition more smooth. As a result, the model can learn a more comprehensive representation of the past. Second, the training of the twin networks can also be considered as a knowledge transfer process, in which the knowledge from the backwards network is transferred to the forwarding network, which is used in the inference phase. This knowledge contains information from the future, which is helpful in the prediction/generation process during the inference stage.

The L2 loss~\cite{serdyuk2017twin}, however, is not as good as adversarial training in terms of the knowledge transfer. However, it acts as a regularisation term~\cite{serdyuk2017twin} for the training objectives and makes the specific token aligned between the forwarding and backwards networks when performing twin net training. We find that the L2 loss and adversarial twin net training scheme are complementary. We demonstrate that adding the adversarial loss and combined with L2 regularisation term in twin net training makes the original ParaCNN generate the best results among all our experimental settings.

%^section{Experimental Evaluation}

\section{Experimental Evaluation}
\subsection{Dataset}
We perform all the experiments on the Stanford Visual Paragraph dataset proposed by~\cite{krause2016paragraphs}. In this dataset, each image contains one paragraph. The training, validation and testing sets contain 14,575, 2487 and 2489 images, respectively. We evaluate the BLEU, METEOR, ROUGE-L and CIDEr scores for the generated paragraphs.

\subsection{Implementation Details}
\subsubsection{\textbf{Paragraph Processing.}}
As our model contains an explicit hierarchical structure of paragraph, sentences and words, we split each paragraph into 6 sentences, which contains up to 30 words. Therefore, our model can model a paragraph with a maximum of 180 words. During training, we add a `$<start>$' token in front of each paragraph. We skip the words which appear less than 2 times during the vocabulary establishment, hence we have a total of 8668 words in our vocabulary.
We then use one-hot encoding to encode each word before passing it to a two-layer trainable word embedding module.
\begin{table}[!t]
    \caption{Parameter Settings: `Projection' means the visual vector after a fully-connected layer for the visual input. `Topic' indicates the dimension of topic vector and `Context' is the dimension of the context vector.}
    \vspace{0.1cm}
    \centering
    \resizebox{\linewidth}{!}{
        \begin{tabular}{|c|c|c|c|c|c|}
            \hline
            \hline
            Vocabulary & Paragraph & Sentence & Max Epochs & Learning Rate & Optimiser  \\
            8668 & 6 sentences & 30 words & 40 & 4e-4 & RMSprop    \\
            \hline
            Visual Input & Projection & Topic & Word Embedding & Context & Convolutional Channels  \\
            4096 & 512 & 512 & 512 & 512 & 512 \\
            \hline
            \hline
        \end{tabular}
    }
    \label{parameters}
\end{table}
\hspace{-0.3cm}

\subsubsection{Training Details.}

To make a fair comparison, the visual representation of each image is the same as that of~\cite{krause2016paragraphs,chatterjee2018diverse,liang2017recurrent}, which is from the dense captioning~\cite{johnson2016densecap} visual features. We map the visual features to a 512 dimension vector by using fully connected layers and max pooling, which may harm the final performance since the dimension of the visual features is only half of that in~\cite{krause2016paragraphs,chatterjee2018diverse,liang2017recurrent}. However, our primary focus is on the feasibility of our model with the only convolutional operation. For the adversarial training of the twin ParaCNN, we use a one-layer bi-GRU as the discriminator, and the dimension of both the hidden states and outputs of the GRU model are set as 512. During the adversarial training, following the practice of W-GAN~\cite{arjovsky2017wasserstein}, we train the discriminator 5 steps and the generator 1 step, alternatively. We use RMSprop optimisation techniques with backpropagation in training. We implement our models with the PyTorch~\cite{paszke2017automatic} deep learning package. All experiments are conducted on a PC with Ubuntu 18.04 system and equipped with an Nvidia Geforce 2080Ti GPU. A detailed configuration of the parameters is shown in Table~\ref{parameters}.

\subsection{Results}

\begin{figure}[t]
    \centering
    \includegraphics[width=0.7\linewidth]{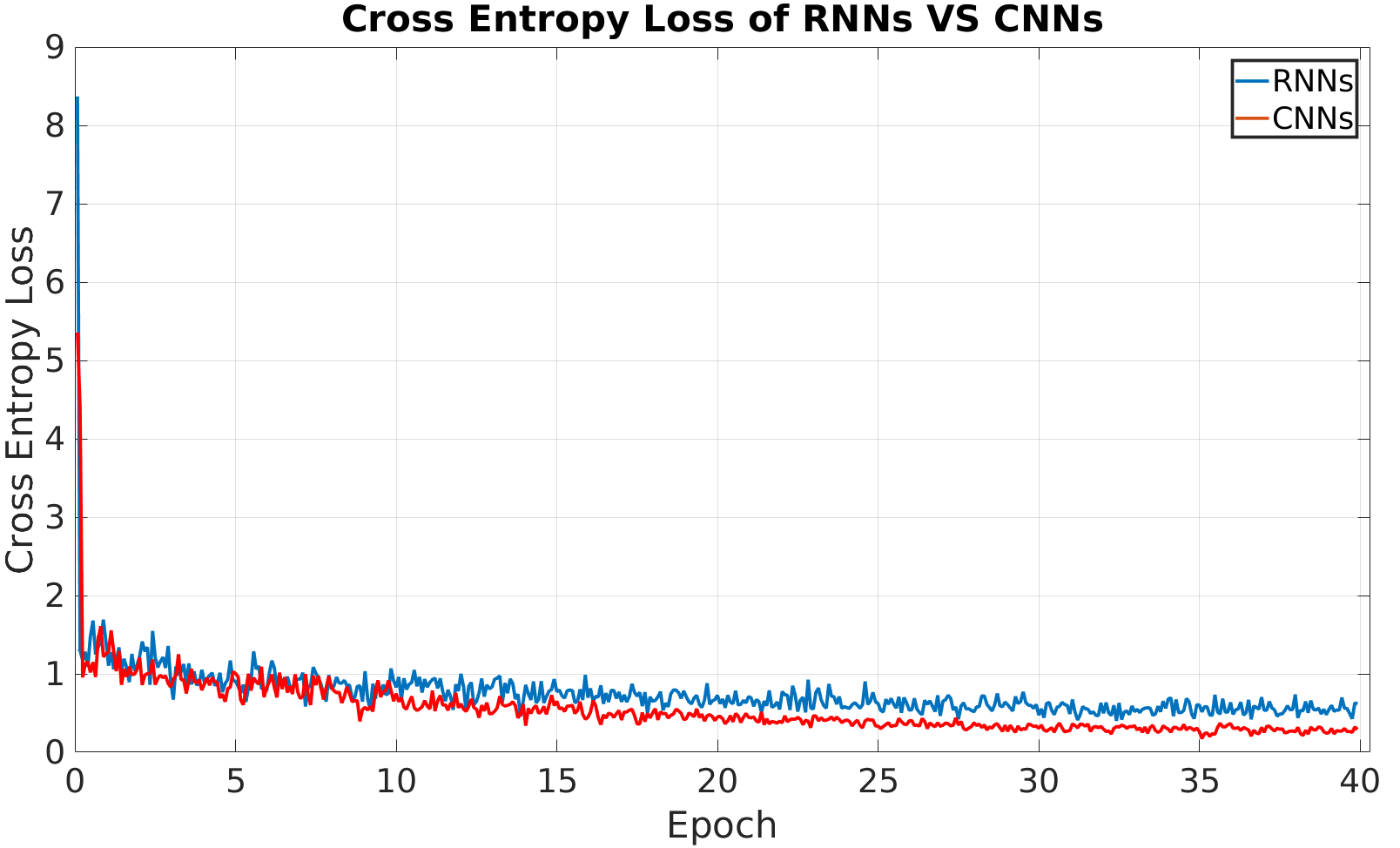}
    \caption{The cross entropy loss of the hierarchical RNNs and the proposed ParaCNN. Our ParaCNN converges faster and has a lower loss value than the hierarchical RNN counterpart.}
    \label{img:loss}
\end{figure}

\begin{table}[!t]
    \caption{The Model Complexity and Time Efficiency of the ParaCNN.}
    \centering
        \begin{tabular}{|c|c|c|}
            \hline
            \hline
            Methods  & Model Complexity & Time Efficiency \\
            \hline
            Hierarchical RNNs & 165.65 M Parameters &  2506s /epoch  \\
            ParaCNN  & 38.27 M Parameters & 2078s /epoch \\
            ParaCNN + Attention  & 38.38 M Parameters  &   2101s /epoch   \\
            \hline
            \hline
        \end{tabular}    \label{model_complexity}
\end{table}

%\subsubsection{Parameter Ablation Studies.}
\subsubsection{\textbf{The Impact of Hyper-parameters.}}
\begin{table*}[!t]
    \caption{Ablation Studies on the Parameters of the ParaCNN: Kernel of 5, depth of 4 for the topic and depth of 5 for the sentence convolutions yields the best performance.}
    \resizebox{\linewidth}{!}{
        \centering
        \begin{tabular}{|c|c|c|c|c|c|c|c|c|}
            \hline
            \hline
            Methods  & BLEU-1 & BLEU-2 & BLEU-3 & BLEU-4 & METEOR & ROUGE-L&  CIDEr \\
            \hline
            Kernels (Topics, Words) = (5, 7), Depth (Topics, Words) = (3, 6)  & 35.6 & 19.1 & 9.9 & 5.1 & 13.5 & 25.3 & 10.3 \\
            Kernels (Topics, Words) = (3, 5), Depth (Topics, Words) = (3, 6) & 37.5 & 20.3  & 10.7  & 5.5 & 14.1 & 25.7 & 12.3  \\
            Kernels (Topics, Words) = (5, 5), Depth (Topics, Words) = (3, 5)  & 39.5 & 22.5 & 13.0 & 7.5 & 15.1 & 28.2 &  15.1 \\
            Kernels (Topics, Words) = (5, 5), Depth (Topics, Words) = (3, 6)  & 40.0 & 22.7 & 13.1 & 7.5 & 15.3 & 28.4 & 15.6 \\
            Kernels (Topics, Words) = (4, 5), Depth (Topics, Words) = (4, 5)  & 39.9 & 22.7 & 13.2 & \textbf{7.6} & 15.2 & \textbf{28.5} & 16.2 \\
            Kernels (Topics, Words) = (5, 5), Depth (Topics, Words) = (4, 5)  & \textbf{40.9} & \textbf{23.3} & \textbf{13.3} & 7.5 & \textbf{15.5} & 28.2 & \textbf{16.4} \\
            \hline
            \hline
        \end{tabular}
    }    \label{parameters_ablation}
\end{table*}
We perform ablation studies on the kernel size and the depth of the topic convolution and word convolution modules, which are shown in Table~\ref{parameters_ablation}.
We set various kernel sizes and depths for the topic convolution and word convolutions. The kernel size of 5 for both the topic and word convolutions yields the best performance. Increasing or decreasing the kernel size of topic convolution does not bring performance increase.
We also set the maximum depth of the whole network as 9, then 4 convolutional layers for the topic and 5 convolutional layers for word generation in our studies.

\subsubsection{\textbf{The Impact of Different Model Configurations for ParaCNN.}}
\begin{table*}[!t]
    \caption{The Performance Comparisons on the Different Scheme of Modules: Overall, our ParaCNN with attention mechanism generates the best results.}
    \resizebox{\linewidth}{!}{
        \centering
        \begin{tabular}{|c|c|c|c|c|c|c|c|c|}
            \hline
            \hline
            Methods  & BLEU-1 & BLEU-2 & BLEU-3 & BLEU-4 & METEOR & ROUGE-L&  CIDEr \\
            \hline
            Topics = RNN, Words = RNN (Results of~\cite{hrnnxinpeng})  &  35.6 &   18.0     &     9.1    &   4.5    &   13.9    &  - & 10.6 \\
            Topics = RNN, Words = CNN & 36.1 & 18.9 & 9.9 & 5.4 & 13.6 & 26.0 & 12.1    \\
            Topics = CNN, Words = CNN (No context) & 38.1 & 21.2 & 12.3 & 7.1 & 14.6 & 28.1 & 14.4 \\
            Topics = CNN, Words = CNN (ParaCNN) & \textbf{40.9} & \textbf{23.3} & {13.3} & {7.5} & {15.5} & {28.2} & {16.4} \\
            Topics = CNN, Words = CNN (ParaCNN, Attention) & 39.3 & 22.8 & \textbf{13.5} & \textbf{8.1} & \textbf{15.6} & \textbf{29.4} & \textbf{17.5} \\
            \hline
            \hline
        \end{tabular}
    }    \label{result_ablation}
\end{table*}
Firstly, the topic and word modules are all set as RNNs, which is implemented by~\cite{hrnnxinpeng}. The result is shown in Table~\ref{result_ablation}, which is not satisfactory.

We then replace the word module to a CNN decoder, and see a slight increase in all the language evaluation metrics, which shows consistency to previous research~\cite{aneja2018convolutional} that convolutional decoder is slightly better than an RNN decoder.

The CNNs are applied for both topic and word module, but without contextual information during the topic convolution. We see a big increase in the all the evaluation metrics, which shows the superiority of the CNNs in paragraph generation, especially in topic modelling since the improvement is largely brought by using the CNNs in the topic generation.

Subsequently, we add contextual information from the last generated sentence to the new one during the topic generation, and also see an obvious improvement in the results, as presented in the last row of Table~\ref{result_ablation}.
%    \end{itemize}
We also visualise the loss value of training of the hierarchical RNNs and our ParaCNN in Figure~\ref{img:loss}, which shows our ParaCNN starts to converge faster than the RNN-based model from the 10th epoch.

To analyse the model complexity and time efficiency of the ParaCNN, we record the model complexity and time efficiency of training of the Hierarchical RNNs and our ParaCNN in Table~\ref{model_complexity}. The recorded numerical results show that our ParaCNN has less model complexity and higher time efficiency in training.

\subsubsection{\textbf{The Impact of the Transformer-based Sentence Generator.}}
We replace the word convolution module with a state-of-the-art transformer-based sentence generator~\cite{cornia2019m}, the results are presented in Table~\ref{transformer}. The results are not as good as our CNN-based module.

\begin{table*}[!t]
    \caption{The performance of the Transformer-based Sentence Generator.}
    \centering
    \resizebox{\linewidth}{!}{
        \begin{tabular}{|c|c|c|c|c|c|c|c|c|}
            \hline
            \hline
            Methods  & BLEU-1 & BLEU-2 & BLEU-3 & BLEU-4 & METEOR & ROUGE-L&  CIDEr \\
            \hline
            ParaCNN & 40.9 & {23.3} & {13.3} & {7.5} & {15.5} & {28.2} & {16.4} \\
            ParaCNN (Transformer) & 39.9 & 22.0 & 12.5 & 7.1 & 15.2 & 27.5 & 14.7 \\
            \hline
            \hline
        \end{tabular}    \label{transformer}
    }
\end{table*}

\subsubsection{\textbf{The Impact of Number of Sentences within a Paragraph.}}

\begin{table}[!t]
    \caption{The Impact of Training Paragraph Length: We increase the training paragraph length to 7 sentences, but see no much differences on the results.}
    \vspace{0.1cm}
    \centering
    \resizebox{\linewidth}{!}{
        \begin{tabular}{|c|c|c|c|c|c|c|c|c|}
            \hline
            \hline
            Methods  & BLEU-1 & BLEU-2 & BLEU-3 & BLEU-4 & METEOR & ROUGE-L&  CIDEr \\
            \hline
            ParaCNN (7 Sent.) & 38.9  & 22.1  & 12.6  &  7.2    & \textbf{16.3}    &  28.2   &  \textbf{16.9}  \\
            ParaCNN (6 Sent.)  & \textbf{40.9} & \textbf{23.3} & \textbf{13.3} & \textbf{7.5} & {15.5} & 28.2 & 16.4 \\
            \hline
            \hline
        \end{tabular}    \label{result_training_num}
    }
\end{table}
\begin{table}[!t]
    \caption{The Impact of Generated Paragraph Length: We use the trained ParaCNN model to generate a various number of sentences, also including the adaptive number achieved by using another standalone model to determine the number of sentences based on the visual inputs.}
    \centering
    \resizebox{\linewidth}{!}{
        \begin{tabular}{|c|c|c|c|c|c|c|c|c|}
            \hline
            \hline
            Methods  & BLEU-1 & BLEU-2 & BLEU-3 & BLEU-4 & METEOR & ROUGE-L&  CIDEr \\
            \hline
            Twin ParaCNN (5 Sent.) & 37.9 & 21.5  & 12.4  &  7.1    & 14.7    &  28.1   &  13.6  \\
            Twin ParaCNN (7 Sent.) & 38.1 & 21.8  & 12.5  &  7.2    & 16.8    &  28.3   &  16.7  \\
            Twin ParaCNN (6 Sent.) & \textbf{41.5} & \textbf{23.6}  & 13.5  &  \textbf{7.7}    & 15.9    &  \textbf{28.4}   &  \textbf{18.6}   \\
            \hline
            Ad. Sent. (Max 6) & 38.0 & 21.7 & 12.4 & 7.1 & 14.7 & 27.9 & 15.9 \\
            \hline
            Ad. Sent. (Min 5, Max 6) & 39.5 & 22.5 & 12.9 & 7.4 & 15.2 & 28.3 & 15.7 \\
            Ad. Sent. (Min 5, Max 7) & 41.0 & 23.3 & 13.4 & 7.7 & 15.5 & 28.3 & 15.7 \\
            Ad. Sent. (Min 6, Max 7) & 40.2 & 22.9 & 13.1 & 7.5 & \textbf{16.2} & \textbf{28.4} & \textbf{18.6} \\
            \hline
            \hline
        \end{tabular}    \label{result_num}
    }
\end{table}

As presented in Table~\ref{result_num}, our ParaCNN can generate a various number of sentences within a paragraph, due to the flexibility of convolutional operations. Although 6 sentences still yield the best result, we show that in real-world applications, our ParaCNN is more flexible to generate longer paragraph to meet the demand. When we re-train the model using 7 sentences, the ParaCNN tends to be more consistent with human consensus since there is an increase in METEOR and CIDEr, as presented in Table~\ref{result_training_num}.

Besides, we train another standalone model to predict the number of generated sentences. This model is a three-layer fully connected neural network, trained using the ground-truth of the number of the sentence in the training set. We notice a slight decrease in the evaluation metrics in this adaptive length inference. When limiting the number of the sentence to a specific range, there is an increase in performance. A suitable length of the generated paragraph is vital in maintaining the performance. Nevertheless, we prove the flexibility of our model that it can generate an arbitrary length of a paragraph.

\subsubsection{\textbf{The Impact of the Visual Attention Mechanism on the ParaCNN.}}
\begin{table*}[!t]
    \caption{The Performance of the Attention Mechanism for the ParaCNN: The visual attention mechanism improves the performance of each scheme.}
    \centering
    \resizebox{\linewidth}{!}{
        \begin{tabular}{|c|c|c|c|c|c|c|c|c|}
            \hline
            \hline
            Methods  & BLEU-1 & BLEU-2 & BLEU-3 & BLEU-4 & METEOR & ROUGE-L&  CIDEr \\
            \hline
            ParaCNN (No Visual Attention) & \textbf{40.9} & \textbf{23.3} & {13.3} & {7.5} & {15.5} & {28.2} & {16.4} \\
            \hline
            ParaCNN (With Visual Attention) & 39.3 & 22.8 & \textbf{13.5} & \textbf{8.1} & \textbf{15.6} & \textbf{29.4} & \textbf{17.5} \\
            \hline
            \hline
        \end{tabular}    \label{result_attention}
    }
\end{table*}

Additionally, the visual attention mechanism~\cite{xu2015show} is applied in the word generation convolutional module. There are 5 convolutional layers in the word generation module, as presented in Table~\ref{parameters_ablation}. We add the visual attention mechanism in the 2nd and the 4th layer of the word generation module instead of every layer. The results are shown in Table~\ref{result_attention}. It can be seen in the table that the visual attention mechanism generally improves the results, since the visual attention mechanism forces the ParaCNN on the more precise location of the visual features, thus generate more accurate concepts.

\subsubsection{\textbf{The Impact of Twin Net Training and Adversarial Optimisation.}}
\begin{table*}[!t]
    \caption{The Performance of Twin Net Training: The twin net training improves most of the metrics. Especially, the adversarial twin net improves the CIDEr metric significantly.}
    \centering
    \resizebox{\linewidth}{!}{
        \begin{tabular}{|c|c|c|c|c|c|c|c|c|}
            \hline
            \hline
            Methods  & BLEU-1 & BLEU-2 & BLEU-3 & BLEU-4 & METEOR & ROUGE-L&  CIDEr \\
            \hline
            ParaCNN & 39.3 & 22.8 & {13.5} & {8.1} & {15.6} & {29.4} & {17.5} \\
            Twin ParaCNN (L2) & 40.9 & {23.8} & {14.2} & {8.5} & {15.9} & {29.9} & 18.3 \\
             Twin ParaCNN (Adversarial) & 41.6 & 24.0  & 14.0  &  8.5    & 16.2    &  29.0   &  {19.1}  \\
             \hline
            Twin ParaCNN (L2, Rep. Penalty Sampling) & {42.5} & {25.3} & {15.3} & {9.2} & {16.4} &  \textbf{30.2} &  19.0    \\
            Twin ParaCNN (Adversarial, Rep. Penalty Sampling) & \textbf{43.2} & \textbf{25.6} & \textbf{15.4} & \textbf{9.5}  & \textbf{16.8} & 29.7  & \textbf{20.5}  \\
            \hline
            \hline
        \end{tabular}    \label{result_twin}
    }

\end{table*}
\begin{table*}[!t]
    \caption{The Results of the L2 Twin Net, Adversarial Twin Net, and the Combination of Them: We see an improvement of using adversarial training on twin net. Furthermore, the combination of the L2 regularisation term and the adversarial twin net training yields the best results.  }
    \centering
    \resizebox{\linewidth}{!}{
        \begin{tabular}{|c|c|c|c|c|c|c|c|c|}
            \hline
            \hline
            Methods  & BLEU-1 & BLEU-2 & BLEU-3 & BLEU-4 & METEOR & ROUGE-L&  CIDEr \\
            \hline
            Twin ParaCNN (L2) & 40.9 & {23.8} & {14.2} & {8.5} & {15.9} & {29.9} & 18.3 \\
            %Twin ParaCNN (L2, Rep. Penalty Sampling) & {42.5} & {25.3} & {15.3} & {9.2} & {16.4} &  {30.2} &  19.0    \\
             Twin ParaCNN (Adversarial) & 41.6 & 24.0  & 14.0  &  8.5    & 16.2    &  29.0   &  {19.1}  \\
            %Twin ParaCNN (Adversarial, Rep. Penalty Sampling) & {43.2} & {25.6} & {15.4} & \textbf{9.5}  & {16.8} & 29.7  & {20.5}  \\
            \hline
            Twin ParaCNN (L2 + Adversarial) & 42.7 & 24.9 & 14.5 & 8.5 & 16.8 & 29.9 & 19.6 \\
            Twin ParaCNN (L2 + Adversarial, Rep. Penalty Sampling) & \textbf{43.3} & \textbf{25.8} & \textbf{15.6} &  \textbf{9.5} & \textbf{17.2} & \textbf{30.3} & \textbf{20.6} \\
            \hline
            \hline
        \end{tabular}    \label{result_combin}
    }
\end{table*}

\begin{table*}[!t]
    \caption{The Impact of the Learning Rate of the Discriminator in Twin Net Training: A suitable learning rate for the discriminator is critical in the adversarial twin net scheme.}
    \centering
    \resizebox{\linewidth}{!}{
        \begin{tabular}{|c|c|c|c|c|c|c|c|c|}
            \hline
            \hline
            Methods  & BLEU-1 & BLEU-2 & BLEU-3 & BLEU-4 & METEOR & ROUGE-L&  CIDEr \\
            \hline
            Twin ParaCNN (Adversarial, lr 3e-4) & 40.7 & 23.5 &13.8 &8.2 &15.9& 29.2 &18.5 \\
            Twin ParaCNN (Adversarial, lr 3e-4, Rep. Penalty Sampling) & 42.4 & 25.1 & 15.0 & 9.0 & 16.4  & \textbf{29.8}  & 19.5 \\
            \hline
             Twin ParaCNN (Adversarial, lr 2e-4) & 41.6 & 24.0  & 14.0  &  8.5    & 16.2    &  29.0   &  {19.1}  \\
            Twin ParaCNN (Adversarial, lr 2e-4, Rep. Penalty Sampling) & \textbf{43.2} & \textbf{25.6} & \textbf{15.4} & \textbf{9.5}  & \textbf{16.8} & 29.7  & \textbf{20.5}  \\
            \hline
            \hline
        \end{tabular}    \label{result_lr}
    }
\end{table*}
We validate the effectiveness of the twin net training scheme on the ParaCNN model by forcing the distance between the hidden features (the last fully-connected features) of the forwarding network and backwards network close. We train the networks from scratch by 40 epochs, and the results are presented in Table~\ref{result_twin}. We see a clear increase in all the language evaluation metrics, which proves that the twin net training is effective for ParaCNN. We then apply the twin net with adversarial training framework to optimise the ParaCNN and the results are presented in Table~\ref{result_twin}. The adversarial training of twin net bring much improvement on CIDEr, which evaluates how much the generated paragraph matches the human consensus. In addition, as Table~\ref{result_lr} reveals, the learning rate of the discriminator is critical for improving the performance. We find improved results of using adversarial twin net training. Moreover, as Table~\ref{result_combin} shows, the L2 loss combined with the adversarial twin net training yields the best result on all the evaluation metrics.

\subsubsection{\textbf{The Impact of the Different Pooling Method in Context Generation.}}
The pooling method in context generation, as shown in Figure~\ref{para} can be set mean pooling, which is simple averaging the information. Therefore, we use a multi-headed self-attention model~\cite{vaswani2017attention} to replace the mean-pooling operation. The self-attention can discover the internal relationship in the previous generated sentence and selects more informative knowledge for the next sentence generation.

\begin{table*}[!t]
    \caption{The Performance Comparison between Different Pooling Method for Context Generation.}
    \centering
    \resizebox{\linewidth}{!}{
        \begin{tabular}{|c|c|c|c|c|c|c|c|c|}
            \hline
            \hline
            Methods  & BLEU-1 & BLEU-2 & BLEU-3 & BLEU-4 & METEOR & ROUGE-L&  CIDEr \\
            \hline
            ParaCNN (Mean Pooling) & 39.3 & 22.8 & {13.5} & {8.1} & {15.6} & {29.4} & {17.5} \\
            ParaCNN (Self-attention Pooling)& 40.6 &  {23.9} & {14.3}  & {8.6} & {16.1} & {29.8} & {18.2}  \\
            \hline
            Twin ParaCNN (Mean Pooling)  & \textbf{42.7} & 24.9 & 14.5 & 8.5 & 16.8 & 29.9 & 19.6 \\
            Twin ParaCNN (Self-attention Pooling) & {42.0} & \textbf{25.0} & \textbf{14.9} & \textbf{8.8} & \textbf{17.0} & \textbf{30.1} &  \textbf{20.4} \\
            \hline
            \hline
        \end{tabular}    \label{transformer}
    }
\end{table*}
\subsubsection{\textbf{Comparison with the State-of-the-art Methods.}}

\begin{table*}[!t]
    \caption{The Performance Comparison with the State-of-the-art Methods on the Visual Paragraph Dataset: We achieve state-of-the-art results on the evaluation metrics.}
    \vspace{0.1cm}
    \resizebox{\linewidth}{!}{
        \renewcommand\arraystretch{1}
        \centering
        \begin{tabular}{|c|c|c|c|c|c|c|c|c|}
            \hline
            \hline
            Category &    Methods  & BLEU-1 & BLEU-2 & BLEU-3 & BLEU-4 & METEOR & CIDEr \\
            \hline
            \multirow{6}{*}{Flat Models} &
            Sentence-Concat~\cite{krause2016paragraphs} & 31.1 & 15.1 & 7.6 & 4.0 & 12.1 & 6.8 \\
            &    Template~\cite{krause2016paragraphs} & 37.5 & 21.0 & 12.0 & 7.4 & 14.3 & 12.2 \\
            &        Image-Flat~\cite{krause2016paragraphs} & 34.0 & 19.1 & 12.2 & 7.7 & 12.8 & 11.1  \\
            &        Top-down Attention~\cite{anderson2018bottom} &32.8 &19.0 &11.4 &6.9 &12.9 &13.7 \\
            &        Self-critical~\cite{melas2018training}  &29.7& 16.5 &9.7 &5.9& 13.6& 13.8 \\
            &        DAM-Att~\cite{wang2018look} &35.0 &20.2 &11.7 &6.6 &13.9 &17.3 \\
            & Meshed Transformer~\cite{cornia2019m} & 37.5 & 22.3 & 13.7 & 8.4 & 15.4 & 16.1 \\
            \hline
            \multirow{3}{*}{Hierarchical Models} &
            Regions-Hierarchical~\cite{krause2016paragraphs} & 41.9 & 24.1 & 14.2 & 8.7 & 16.0 & 13.5  \\
            &        RTT-GAN~\cite{liang2017recurrent} & 42.0 & 24.9 & {14.9} & 9.0 &  17.1 & 16.9  \\
            &        Diverse (VAE)~\cite{chatterjee2018diverse} &  42.4  & {25.6} & 15.2 & {9.4} & \textbf{18.6} & {20.9} \\
            &   CAVP~\cite{zha2019context}  & 42.0 & 25.9 & 15.3& 9.3 & 16.8 &  {21.1}   \\

            \hline
            %\multirow{3}{*}{~\cite{polis2019paragraph}}
        %    &              Image-Flat~\cite{polis2019paragraph}  &  37.7 &21.9& 12.8& 7.4  & 15.0 & 17.8 \\
        %    &         Regions-Hierarchical~\cite{polis2019paragraph} & 40.1 &22.2& 12.3 & 6.8 &15.1 & 17.0 \\
        %    &         Diverse (VAE)~\cite{polis2019paragraph}  & 41.1 & 23.2 & 13.2 & 7.5 & 15.6 & 16.3 \\

            \hline
            \multirow{6}{*}{Ours}% & Twin ParaCNN (Adversarial) & 41.6 & 24.0  & 14.0  &  8.5    & 16.2     &  {19.1}  \\
            %&    Twin ParaCNN (Adversarial, Rep. Penalty Sampling) & {43.2} & {25.6} & {15.4} & \textbf{9.5}  & {16.8}  & {20.5}  \\
               &     Twin ParaCNN (Mean Pooling) & 42.7 & 24.9 & 14.5 & 8.5 & 16.8 & 19.6 \\
            & Twin ParaCNN (Mean pooling, Rep. Penalty Sampling) & \textbf{43.3} & \textbf{25.8} & \textbf{15.6} &  \textbf{9.5} & 17.2 & 20.6 \\
            \cline{2-8}
               &     Twin ParaCNN (Self-attention Pooling) & {42.0} & {25.0} & {14.9} & {8.8} & {17.0}  &  {20.4} \\
            & Twin ParaCNN (Self-attention Pooling, Rep. Penalty Sampling) & 42.9 & 25.6 & \textbf{15.6}  & 9.3 &  18.0 & \textbf{21.6}  \\
            \hline
            
            Human~\cite{krause2016paragraphs} & Annotation  &      42.9 & 25.7 & 15.6 & 9.7  & 19.2  & 28.6 \\
            \hline
            \hline
        \end{tabular}
    }
    \label{result}
\end{table*}%
\textbf{(a) Flat models:} \textbf{(a1) Sentence-Concat.} Two sentence-level captioning models which are
\textbf{Neural talk}~\cite{karpathy2015deep} and \textbf{NIC}~\cite{vinyals2015show} pre-trained on the MSCOCO~\cite{lin2014microsoft} dataset are adopted to predict 5 sentences for each image.
\textbf{(a2) Image-Flat.} Image-Flat~\cite{krause2016paragraphs} applied a deep CNN to encode the image and train a flat RNN model to generate the entire paragraph.
\textbf{(a3) DAM-Att.} DAM-Att~\cite{wang2018look} used additional information of depth image to enhance the spatial relationship information. Our model is a hierarchical CNN model which is less similar to the flat RNNs. Also, our ParaCNN tends to perform much better than the flat models.
\textbf{(b) Hierarchical Models:} \textbf{(b1) Region-Hierarchical.} Region-Hierarchical~\cite{krause2016paragraphs} applied hierarchical RNNs to model the hierarchy from words to paragraph. The ParaCNN generates much higher results on the evaluation metrics, as presented in Table~\ref{result_ablation}. Besides, our ParaCNN converges faster than the hierarchical RNNs model.
\textbf{(b2) RTT-GAN.} RTT-GAN~\cite{liang2017recurrent} realised the hierarchical RNNs in a GAN setting where the generator mimics the human-annotated paragraphs and tries to fool the discriminator.
\textbf{(b3) Diverse(VAE).} Diverse(VAE)~\cite{chatterjee2018diverse} modelled the paragraph distribution with variational auto-encoder (VAE), which can preserve the consistency of the global topics of paragraphs. This method focuses on the diversity of the topic in the dataset and the internal consistency within one paragraph. In contrast, the proposed ParaCNN focuses on the feasibility of the pure convolutional operations on the visual paragraph generation and the effectiveness of the twin net training scheme. \textbf{(b4) Context-Aware Visual Policy network (CAVP).} CAVP~\cite{zha2019context} proposed a reinforcement learning scheme by considering the previous visual attentions as context and deciding whether the context is used for the current word or sentence generation. The focus of CAVP~\cite{zha2019context} and our work are different, but we also consider the context information of the previously generated sentence and see the effectiveness of the contextual information.
%%~\cite{polis2019paragraph} comprehensively studied the current mainstream methods on the visual paragraph generation and re-implement each of them. We outperform the re-implementations, as presented in Table~\ref{result}. Hence, we conclude that our ParaCNN model is comparative to the current state-of-the-art RNN-based methods, and the adversarial twin net training is effective for the ParaCNN.

\begin{figure*}[t]
    \centering
    \includegraphics[width=\linewidth]{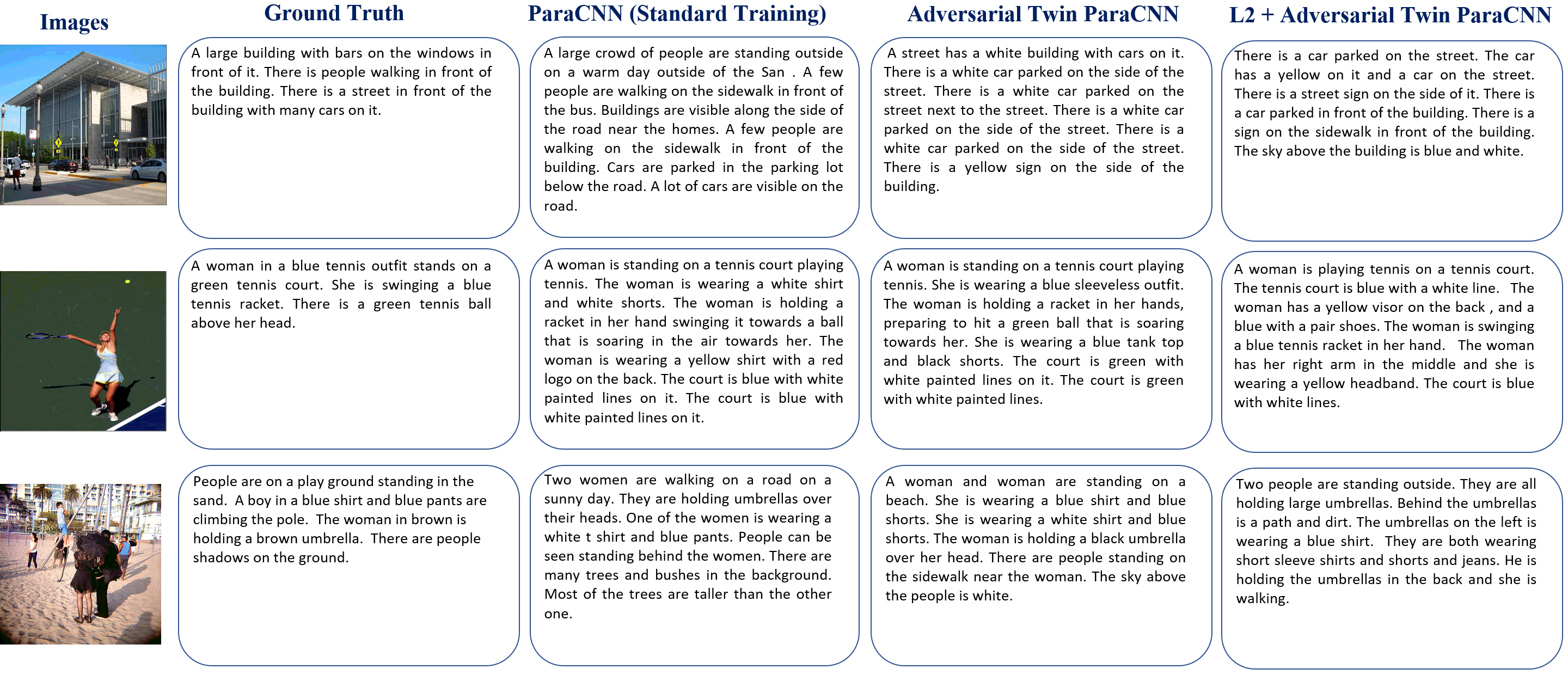}
    \caption{The visualisation of images and corresponding generated paragraphs from different methods. The figure shows that the ParaCNN can generate more detailed paragraph, and the adversarial twin net training indeed improves the ParaCNN baseline by generating more accurate sentences. More interestingly, the ParaCNN with twin net generates more novel concepts of subtle objects and their relationships, which are neglected in other methods, even in the ground truth. The best results are generated with the combination of L2 loss and adversarial twin net training.}
    \label{img:vis}
    \vspace{-0.5cm}
\end{figure*}

\subsubsection{\textbf{Visualisation of the Generated Paragraphs.}}
As shown in Figure~\ref{img:vis}, we present the generated paragraphs from the ParaCNN with the standard training, adversarial twin net training and the L2 combined with twin net ParaCNN. The twin net training scheme generates much different paragraph from the ParaCNN baseline and includes more fine-grained details, which proves that it indeed improves the ParaCNN training.

\section{Conclusions}
This paper studies the task of visual paragraph generation, which could describe images with long paragraph thus facilitate the real-world applications such as image retrieval, automatic navigation and disabled support. Specifically, we propose a novel model, called ParaCNN, by only using convolutional operations on the multimodal data of image and natural language. The ParaCNN can form a hierarchical structure on the word, sentence and paragraph, thus solving the long sequence generation problem at the same time. The ParaCNN is flexible to generate an arbitrary number of sentences, which is more suitable for real-world applications. Moreover, we propose an adversarial twin net training scheme to enable the ParaCNN to contain backwards information, which improves the performance. We conduct extensive experiments to validate the proposed method and achieve state-of-the-art results.

\clearpage
\bibliographystyle{splncs04}
\bibliography{egbib}
\end{document}